\documentclass[sn-mathphys,Numbered,iicol]{sn-jnl}

\usepackage[finalnew]{trackchanges}

\usepackage{graphicx}%
\usepackage{subcaption}
\usepackage{multirow}%
\usepackage{amsmath,amssymb,amsfonts}%
\usepackage{amsthm}%
\usepackage{mathrsfs}%
\usepackage[title]{appendix}%
\usepackage{xcolor}%
\usepackage{textcomp}%
\usepackage{manyfoot}%
\usepackage{booktabs}%
\usepackage{algorithm}%
\usepackage{algorithmicx}%
\usepackage{algpseudocode}%
\usepackage{listings}%
\usepackage{xcolor}
\definecolor{dark_green}{RGB}{9, 106, 9}
\newcommand{\mysection}[1]{\vspace{2pt}\noindent\textbf{#1}}

\usepackage{xcolor, color, colortbl}         
\definecolor{Highlight}{HTML}{39b54a}  

\usepackage{amssymb}
\usepackage{pifont}
\usepackage{algorithm}
\usepackage{listings}
\usepackage{etoolbox}
\makeatletter
\AfterEndEnvironment{algorithm}{\let\@algcomment\relax}
\AtEndEnvironment{algorithm}{\kern2pt\hrule\relax\vskip3pt\@algcomment}
\let\@algcomment\relax
\newcommand\algcomment[1]{\def\@algcomment{\footnotesize#1}}
\renewcommand\fs@ruled{\def\@fs@cfont{\bfseries}\let\@fs@capt\floatc@ruled
  \def\@fs@pre{\hrule height.8pt depth0pt \kern2pt}%
  \def\@fs@post{}%
  \def\@fs@mid{\kern2pt\hrule\kern2pt}%
  \let\@fs@iftopcapt\iftrue}
\makeatother
\newcommand{\cmmnt}[1]{}

\usepackage[normalem]{ulem}

\usepackage{listings}
\usepackage{color}
\definecolor{codegreen}{rgb}{0,0.6,0}
\definecolor{codegray}{rgb}{0.5,0.5,0.5}
\definecolor{codepurple}{rgb}{0.58,0,0.82}
\definecolor{backcolour}{rgb}{1,1,1}
 
\lstdefinestyle{mystyle}{
    backgroundcolor=\color{backcolour},   
    commentstyle=\color{codegreen},
    keywordstyle=\color{magenta},
    numberstyle=\tiny\color{codegray},
    stringstyle=\color{codepurple},
    basicstyle=\footnotesize,
    breakatwhitespace=false,         
    breaklines=true,                 
    captionpos=b,                    
    keepspaces=true,                 
    numbers=left,                    
    numbersep=5pt,                  
    showspaces=false,                
    showstringspaces=false,
    showtabs=false,                  
    tabsize=2
}
 
\makeatletter

\newcommand{\Rmnum}[1]{\expandafter\@slowromancap\romannumeral #1@}
\makeatother

\usepackage{xspace}
\newcommand{\etal}{\emph{et al.}\xspace}
\newcommand{\ie}{\emph{i.e.}\xspace}

\newcommand{\Eg}{\emph{E.g.}\xspace}

\newcommand{\soccer}{football}

\newcommand{\point}{\,.}

\newcommand{\argmax}{\mathop{\mathrm{argmax}}}

\begin{document}

\title[Article Title]{Towards AI-Powered Video Assistant Referee System (VARS) for Association Football}


\author*[1]{\fnm{Jan} \sur{Held}}\email{jan.held@liege.be}

\author*[1,2]{\fnm{Anthony} \sur{Cioppa}}\email{anthony.cioppa@uliege.be}

\author*[2]{\fnm{Silvio} \sur{Giancola}}\email{silvio.giancola@kaust.edu.sa}

\author[3]{\fnm{Abdullah} \sur{Hamdi}}

\author[1]{\fnm{Christel} \sur{Devue}}

\author[2]{\fnm{Bernard} \sur{Ghanem}}

\author[1]{\fnm{Marc} \sur{Van Droogenbroeck}}

\affil[1]{\orgname{University of Liege (ULi{\`e}ge), Belgium}}

\affil[2]{\orgname{King Abdullah University of Science and Technology (KAUST), Saudi Arabia}}

\affil[3]{\orgname{University of Oxford, United Kingdom}}


\abstract{
Over the past decade, the technology used by referees in football has improved substantially, enhancing the fairness and accuracy of decisions.
This progress has culminated in the implementation of the Video Assistant Referee (VAR), an innovation that enables backstage referees to review incidents on the pitch from multiple points of view.
However, the VAR is currently limited to professional leagues due to its expensive infrastructure and the lack of referees worldwide.
\remove{Therefore} \add{In this paper}, we present the 
Video Assistant Referee System (VARS) that leverages the latest findings in multi-view video analysis.
\remove{Specifically, our system provides an automated analysis by recognizing the type of foul and its appropriate sanction. 
VARS sets a new state-of-the-art on the \emph{SoccerNet-MVFoul} dataset, a multi-view video dataset of football fouls.}  
\add{Our VARS achieves a new state-of-the-art on the \emph{SoccerNet-MVFoul} dataset by recognizing the type of foul in $50\%$ of instances and the appropriate sanction in $46\%$ of cases.}
Finally, we conducted a \add{comparative} study \add{to} investigate human performance in classifying fouls and their corresponding severity \add{and compared these findings to our VARS.}
\remove{This study highlights} \add{The results of our study highlight} the potential of our VARS to reach human performance and support football refereeing across all levels of professional and amateur federations.
}




\maketitle

\section{Introduction}
In recent years, technology has played an increasing role in \soccer, revolutionizing how the game is played, coached, and officiated. This transformation extends into the domain of sports video analysis, which encompasses a diverse range of challenging tasks, including player detection and tracking~\cite{Cioppa2020Multimodal, Maglo2022Efficient, Vandeghen2022SemiSupervised, Somers2024SoccerNetGameState}, spotting actions in untrimmed videos~\cite{Cioppa2020AContextaware,Giancola2021Temporally,Hong2022Spotting,Soares2022Temporally, Soares2022Action-arxiv, Giancola2023Towards, Cabado2024Beyond, Kassab_2024}, pass feasibility and prediction~\cite{ArbuesSanguesa2020Using, Honda2022Pass}, summarizing~\cite{Gautam_2022, Midoglu_2024, Sushant_2022, Midoglu2022MMSys}, camera calibration\cite{Magera2024AUniversal}, player re-identification in occluded scenarios~\cite{Somers2023Body}, or dense video captioning for football broadcasts commentaries~\cite{Mkhallati2023SoccerNetCaption, Andrews2024AiCommentator}. Solving these tasks has been taken to a higher level thanks to the emergence of deep learning techniques~\cite{Su2015Multiview, Bahdanau2014Neural-arxiv, Vaswani2017Attention-arxiv}. 
Similar to many other fields in which deep learning has been used, the advancements in sports video understanding heavily rely on the availability of large-scale datasets~\cite{Pappalardo2019Apublic,Yu2018Comprehensive,Scott2022SoccerTrack,Jiang2020SoccerDB,VanZandycke2022DeepSportradarv1}. 
SoccerNet~\cite{Giancola2018SoccerNet,Deliege2021SoccerNetv2,Cioppa2022Scaling, Cioppa2022SoccerNetTracking, Held2023VARS, Cioppa2023SoccerNetChallenge-arxiv, Leduc2024SoccerNetDepth, Held2024XVARS, Gautam2024SoccerNetEchoes-arxiv} stands among the largest and most comprehensive sports dataset, with extensive annotations for video understanding in football.

In refereeing, the biggest \remove{breakthrough} \add{revolution} was introduced by the Video Assistant Referee (VAR) in 2016~\cite{Spitz_2020}. 
The system involves a team of referees located in a video operation room outside the stadium.
These referees have access to all available camera views and check all decisions taken by the on-field referee. 
If the VAR indicates a probable ``clear and obvious error'' (\Eg when the referee misses a penalty or a red card, gives a yellow card to the wrong player, etc.), it will be communicated to the on-field referee who can then review his decision in the referee review area before taking a final decision. 
The VAR helps to ensure greater fairness in the game by reducing the impact of incorrect decisions on the outcome of games. Notably, in 8\% of the matches, the VAR has a decisive impact on the result of the game \cite{DeDiosCrespo2021TheContribution} and it slightly reduces the unconscious bias of referees towards home teams \cite{Holder2021Monitoring}. On average, away teams now score more goals and receive fewer yellow cards\remove{ than before}~\cite{Dufner2023TheIntroduction}. 
Controversial referee mistakes like the famous ``hand of God'' goal by Diego Maradona during the quarter-final match Argentina versus England of the $1986$ FIFA World Cup, 
Josip \v{S}imuni\'{c} getting three yellow cards in a single game at the $2006$ FIFA World Cup, or Thierry Henry's handball preventing the Republic of Ireland from qualifying for the World Cup could have been avoided with the VAR and would have changed football history.
\remove{Furthermore, the VAR has profoundly impacted the mental health of referees. Referees experience less stress and anxiety knowing that the VAR can intervene and that they can correct their mistakes} 

Despite its potential benefits, the use of the VAR technology remains limited to professional leagues. 
The infrastructure of the VAR is expensive, including multiple cameras to analyze the incident from different angles, video operation rooms in various locations, and VAR officials hired to analyze the footage.
Leagues with financial limitations cannot afford the necessary infrastructure to operate the VAR.
In addition to the upfront costs of the infrastructure, there is also an ongoing expense associated with using the VAR. 
The officials who serve as Video Assistant Referees require specialized training \cite{Armenteros2021Educating} and monetary compensation following each game.
\remove{Especially, the officials who serve as Video Assistant Referees must receive specialized training and need to be compensated after each game.}
\add{Given the implementation and operational costs of VAR, its use is currently restricted to professional leagues.}
\remove{Given these costs, it is currently only feasible for professional leagues to implement and operate the VAR, which limits its use.}
\remove{Another} \add{A further} obstacle is the shortage of referees worldwide. 
In Germany, there were only $50{,}241$ active referees during the 2020/2021 season, whereas the number of games played each weekend was around $90{,}000$~\cite{DFB2022Anzahl,Zeppenfeld2023Anzahl}.
The introduction of the VAR requires an additional team of referees per game, which is not feasible for semi-professional or amateur leagues.
\add{Finally, each referee interprets the Laws of the Game}~\cite{IFAB2022Laws} \add{slightly differently, resulting in different decisions for similar actions. Given that the video assistant referee (VAR) changes from one game to another, inconsistencies may arise, with the VAR making different decisions for similar actions across different matches.}

In this paper, we present the 
``Video Assistant Referee System'' (VARS), which could support or extend the current VAR.
Our VARS fulfills the same objectives and tasks as the VAR.
By analyzing fouls from a single or a multi-camera video feed, it indicates a probable ``clear and obvious error'', and can communicate this information to the referee, who will then decide whether to initiate a ``review''.
The proposed VARS automatically analyzes potential incidents that can then be shown to the referee in the referee review area. 
Just like the regular VAR, our VARS serves as a support system for the referee and only alerts him in the case of potential game-changing mistakes, but the final decision remains in the hands of the main referee. 
The main benefit of our VARS is that it no longer requires additional referees, making it the perfect tool for leagues that do not have enough financial or human resources. 

\mysection{Contributions.} We summarize our contributions and novelties as follows:
\textbf{(i)} We propose an upgraded version of the \textit{VARS} presented by Held \etal~\cite{Held2023VARS}. We introduce an attention mechanism on the different views and calculate an importance score to allocate more attention to more informative views before aggregating the views.
\textbf{(ii)} We present a thorough study on the influence of using multiple views and different types of camera views on the performance of our VARS.
\textbf{(iii)} We present a comprehensive human \remove{survey} \add{study} where we compare the performance of human referees, football players, and our VARS on the task of type of foul classification and offense severity classification. \add{Our human study also illustrates the subjectivity of refereeing decisions by examining the inter-rater agreement among referees.}

\section{Methodology}\label{sec:methodology}

We propose an upgraded version of the Video Assistant Referee System, which adds an advanced pooling technique to combine the information from multiple views, extracting the most relevant information based on our \remove{attentional} \add{attention} mechanisms. 
\begin{figure}[t]
    \centering
    \includegraphics[width=\linewidth]{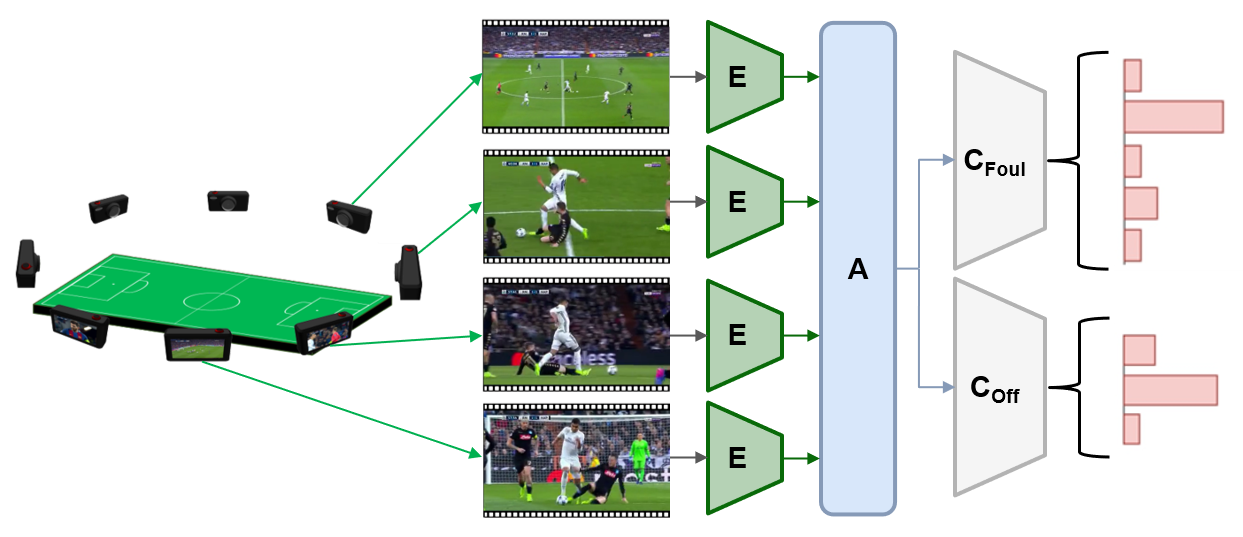}
    \caption{
    \textbf{Architecture of our Video Assistant Referee System.}
    From multi-view video clips input, our system
    \textcolor{dark_green}{encodes per-view video features ($\mathbf{E}$)},
    \textcolor{blue}{aggregates the view features ($\mathbf{A}$)}, and
    \textcolor{black}{classifies different properties ($\mathbf{C_{Foul}}$ and $\mathbf{C_{Off}}$)}. 
    \add{The figure is inspired by} \cite{Held2023VARS}.
    } 
    \label{fig:pipeline}
\end{figure}
The architecture is 
shown in Figure~\ref{fig:pipeline}.
Formally, the VARS takes multiple video clips $\mathbf{v} = \{v_i\}_{1}^n$ as input. Each video clip shows the same action from $n$ different perspectives. 
Each clip $v_i$ is fed into a video encoder $\mathbf{E}$ to extract a  spatio-temporal feature vector $f_i$ of dimension $d$ for each clip $v_i$.
All feature vectors $f_i$ are then stored in a matrix $\mathbf{f}$ as follows:
\begin{equation}\label{equ:input}
\mathbf{f} = 
    \begin{bmatrix}
    f_1, 
    f_2,
    ...,
    f_n
    \end{bmatrix}^T \point
\end{equation}
An aggregation block $\mathbf{A}$ takes $\mathbf{f}$ as input and outputs a single multi-view representation $\mathbf{R}$.
A multi-head classifier, $\mathbf{C}^\text{foul}$ and $\mathbf{C}^\text{off}$, simultaneously predicts the fine-grained type of foul class and the offense severity class.
For each task, the VARS selects the value with the highest confidence  from the respective confidence vector as the final prediction, following:
\begin{equation}
    \mathbf{VARS}^{t} \leftarrow \argmax \mathbf{C}^t_{\theta_{C^t}}(\mathbf{R}),   \forall t \in \{\text{foul}, \text{off}\},
\end{equation}
where $\theta_{C^t}$ corresponds to the parameters of the classification head for task $t$ $\in \{\text{foul}, \text{off}\}$.
The model is trained by minimizing the unweighted summation of both task losses $\mathcal{L}^\text{foul}$ and $\mathcal{L}^\text{off}$.

 \begin{figure*}[t]
 \centering
   \includegraphics[width=1\linewidth]{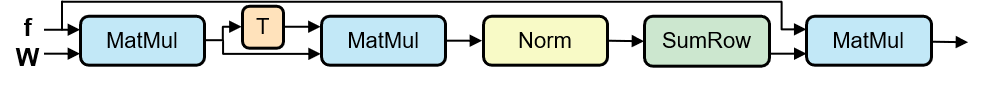}
             \caption{\textbf{Architecture of the attention block.} 
                ``MatMul'' represents matrix multiplication, ``T'' denotes transpose, ``Norm'' signifies normalization, and ``SumRow'' indicates the process of summing each row.} 
             \label{fig:attention_mechanism} 
 \end{figure*}

\mysection{Video Encoder E.}
Based on the work presented in~\cite{Held2023VARS}, the best performance is obtained with a video encoder that extracts spatial and temporal features. 
In the following, we use the state-of-the-art video encoder MViT~\cite{Fan2021Multiscale,Li2022MViTv2} pretrained on Kinetics~\cite{Kay2017TheKinetics-arxiv}, which incorporates a transformer-based architecture with a multiscale feature representation, allowing it to capture spatial and temporal information from video clips. 

\mysection{Multi-view aggregation block A.}
The original paper~\cite{Held2023VARS} used simple mean or max pooling operations to gather the multi-view information into a unique representation. A major drawback of these pooling approaches is that the combination of the feature vectors is fixed and ignores the relationship between the views.
Instead, we propose a new aggregation technique based on an attention mechanism to model such relationships.

Our approach is inspired by the ``Integrating Block'' presented in~\cite{Yang2019Learning}, where each view is associated with an attention score.
However, instead of aggregating multi-view images, we extend the operation to multi-view videos.
Technically, we assign an attention score to each view and then calculate the final representation by a weighted sum of the feature vectors.
There exist several strategies to assign an attention score to a view. A first naive approach consists of passing each feature vector individually into a learned function.
However, this would neglect the relationships between the views and would not provide a relative attention score of the views.
A better approach consists of determining the attention score of each view based on its relationships with the other views.
To do so, we first take the dot product (denoted by $\cdot$) of \textbf{f} multiplied by a matrix $W\in$ $\mathbb{R}^{dxd}$ of trainable weights and its transpose:

\begin{equation}
    \mathbf{S}= \mathbf{f}W \cdot (\mathbf{f}W)^T {.}
\end{equation}
By multiplying the matrix \textbf{f} with its transpose, we compute the dot product between each pair of feature vectors, which measures the similarity between two vectors.
The obtained symmetric similarity matrix \textbf{S} is of dimension $n\times n$, where the value at row $i$ and column $j$ corresponds to the similarity score between view $i$ and view $j$. A higher score indicates a higher similarity between the vectors, while a lower score suggests a lower similarity.
Next, we normalize the similarity scores to get a probability-like distribution, by passing the matrix \textbf{S} through a \textit{ReLU} layer and divide it by the sum of the matrix \textbf{S}, following: 

\begin{equation}
    \mathbf{N}=\frac{\mathit{ReLU(\mathbf{S})}}{\sum_{i=1}^n\sum_{j=1}^n \mathit{ReLU}(\mathbf{S}_{i,j})} \, .
\end{equation}
To obtain the attention score for each view, we sum the values in each row of the normalized similarity matrix \textbf{N}. The attention score for a view $i$ represents the sum of its normalized similarity scores with all other views.
By summing the values in each row of the normalized similarity matrix \textbf{N}, we aggregate the normalized similarity scores for each view. This aggregation reflects how similar a particular view is to all other views collectively. Consequently, the resulting attention score captures a view's overall relevance within the set of views.
The reasoning behind this approach is that if a view is highly similar to many other views, it is considered important because it shares visual content with multiple views. On the other hand, if a view is dissimilar to other views, it might be considered less important since it does not contribute significantly to the collective visual information.
Formally, we take the sum per row to obtain the attention score \textbf{A} per view:

 \begin{equation}
    \mathbf{A}=\sum_{i=1}^n \mathbf{N}_{i,j} \, ,
\end{equation}
where \textbf{A} is a vector of size $n$, where the value $j$ corresponds to the attention score of the view $j$ regarding all other views and itself.
The final representation is given by the sum of the extracted feature vector weighted by their calculated attention score, following:

\begin{equation}
    \mathbf{R}_i= \sum_{j=1}^n \textbf{f}_{i,j} \times \mathbf{A}_j \, .
\end{equation}

\mysection{Classification heads C.}
%
A multi-task classification approach is used to classify simultaneously the type of foul, whether it is an offense or not, and its severity. As both tasks are related, learning them together can lead to improved generalization and a better understanding of each task. The model can leverage the relationships between the two tasks to make better predictions.
Each classification head consists of two dense layers and takes as input the aggregated representation. The output is a vector whose dimensions correspond to the number of classes in each of the classification problems.

\section{Experiments}\label{sec:Experiments}

\subsection{Experimental setup}

\mysection{Tasks.} \label{ExperimentTask}
We test our VARS on the two classification tasks introduced by the SoccerNet-MVFouls dataset~\cite{Held2023VARS}: \textit{Fine-grained foul classification}, which is the task of classifying a foul into one of $8$ fine-grained foul classes (\ie, ``Standing tackling'', ``Tackling'', ``High leg'', ``Pushing'', ``Holding'', ``Elbowing'', ``Challenge'', and ``Dive/Simulation''), and \textit{Offence severity classification}, which is the task of classifying whether an action is an offence, as well as the severity of the foul, defined by four classes: ``No offence'', ``Offence + No card'', ``Offence + Yellow card'', and ``Offence + Red card''.

\mysection{Data.}
\add{The SoccerNet-MVFoul dataset contains $3{,}901$ actions, composed of at least two videos, the live action and at least one replay, see Figure}~\ref{fig:pipeline}. 
\add{The views were manually synchronized by a human and no pre-processing of the video clips is necessary.
Our VARS is trained on clips of $16$ frames, mostly 8 frames before the foul and 8 after the foul, spanning one second temporally with a spatial dimension re-scaled to $224\times224$ pixels.
This approach was chosen because of the high computational cost associated with using a larger number of frames.
Future research could explore whether an increase in frame rate or a larger temporal context enhances performance.}

\mysection{Training details.} 
\remove{Our VARS is trained on clips of $16$ frames, mostly 8 frames before the foul and 8 after the foul, spanning one second temporally with a spatial dimension rescaled to $224\times224$ pixels.}
The encoder \textbf{E} is pre-trained as detailed in the methodology, and the classifier \textbf{C} is trained from scratch, both being trained in an end-to-end fashion. We use a cross-entropy loss, optimized with Adam on a batch size of $6$ samples. The learning rate starts at $5e^{-5}$ and is multiplied by $0.3$ every $3$ steps. 
\add{To artificially increase the dataset size, we use data augmentation and a random temporal shift to have a flexible number of frames used before and after the foul frame annotation during training.}
The model begins to overfit after $7$ epochs and requires approximately $8$ hours of training time on a single NVIDIA V100 GPU.

\mysection{Evaluations metrics.} 
To evaluate the performance of the VARS, SoccerNet-MVFouls uses the classification accuracy, which is the ratio of correctly classified actions regarding the total number of actions.
As SoccerNet-MVFouls~\cite{Held2023VARS} is unbalanced, the authors also suggest a balanced accuracy, which is defined as follows:
\begin{equation}
    \textrm{\mbox{Balanced Accuracy (BA)}} = \frac{1}{N}\sum_{i=1}^{N} \frac{TP_i}{P_i} \, ,
\end{equation}
with \textit{N} being the number of classes, $TP_{i}$ the number of True Positives and $P_{i}$ the number of Positives for class $i$.
To ensure a fair comparison, we use the same training, validation, and test sets as those used in the original paper~\cite{Held2023VARS}.

\subsection{Main results}

Table~\ref{tab:mainResults} shows the results obtained for the fine-grained foul and the offense severity classification task.
Compared to the fixed combination of the feature vectors (mean or max pooling), our novel attention mechanism enhances the model’s ability to identify and classify the type of foul by $5\%$ and the balanced accuracy by $1\%$.
This demonstrates the effectiveness of combining the feature vectors of the different views based on their importance compared to max or mean pooling.
Similarly, the attention mechanism improves the model's performance to determine if an action is a foul and the corresponding severity by $3\%$ and the balanced accuracy remains the same compared to max pooling. One might argue that the performance increase is based on the supplementary parameters introduced by the attention mechanism. However, the attention mechanism only adds an extra 0.1\% of parameters to the model compared to when using max and mean pooling. This suggests that the performance increase derives from the use of the attention mechanism rather than the introduction of additional parameters.

\begin{table}[t]
    \centering
    \resizebox{\linewidth}{!}{
    \begin{tabular}{lc|lc|lc}
        \multicolumn{2}{c|}{} &  \multicolumn{2}{c|}{\bf Type of Foul} & 
        \multicolumn{2}{c}{\bf Offence Severity} \\ \midrule
       \bf Feat. extr. & \bf Pooling  & \bf Acc. & \bf BA. & \bf  Acc. & \bf BA.\\ \midrule
       ResNet & Mean     &0.30     & 0.27  &0.34     &0.25 \\ 
       ResNet  & Max      &0.32  &0.27   & 0.32  &0.24 \\
       R(2+1)D & Mean     &0.32     & 0.34  &0.34     &0.30 \\ 
       R(2+1)D  & Max      &0.34  &0.33   & 0.39  &0.31 \\
       MViT  & Mean     &0.44     & 0.40   &0.38     &0.31 \\ 
       MViT  & Max      &0.45  &0.39   & 0.43  &\textbf{0.34} \\\midrule
       MViT &Attention &\textbf{0.50}  & \textbf{0.41}   &\textbf{0.46} &\textbf{0.34}
    \end{tabular}
    }
    \caption{\textbf{Multi-task classification.} Attention pooling sets a new benchmark on the \emph{SoccerNet-MVFoul} dataset for all the evaluation metrics and tasks. \remove{The} Type of foul classification accuracy increased by $5$\% while the balanced accuracy (BA) increased by $1$\%. We have an increment of $3$\% for the offense severity classification, while the balanced accuracy stays the same.
    }
    \label{tab:mainResults}
\end{table}

\subsection{Detailed analysis}

\begin{figure}[ht]
    \centering
    \includegraphics[width=1\linewidth]{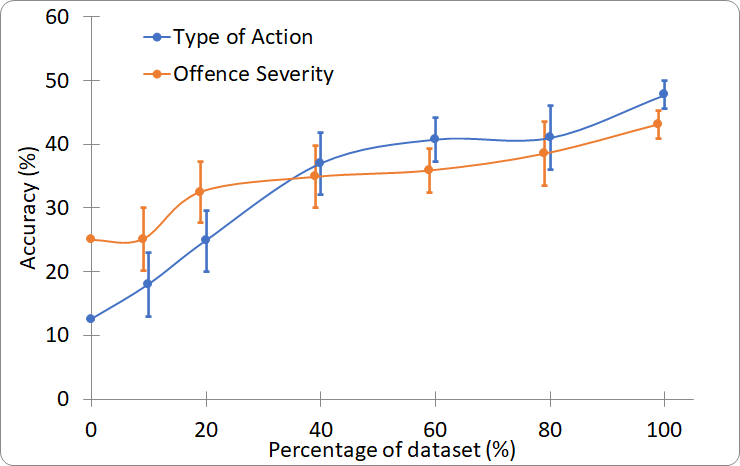}
    \caption{\textbf{Performance evaluation for different dataset sizes.} 100\% of the dataset corresponds to $2{,}319$ actions. For each dataset size, we independently trained and tested the model $10$ times. The tests were all performed on the same test set. The error bar corresponds to the standard deviation. 
    For 0\% of the dataset, we indicate the accuracy by taking a random decision.
    }
    \label{fig:sensitivityAnalysis}
\end{figure}

\mysection{Sensitivity analysis.}
We first investigate the impact of the training dataset size on the performance of our two classification tasks. Figure~\ref{fig:sensitivityAnalysis} shows the evolution of the accuracy regarding different training dataset sizes. 
For each dataset size, we independently trained and tested the model $10$ times to avoid any epistemic uncertainty bias. The
tests were all performed on the same test set. 
As expected, we observe that increasing the dataset size improves the accuracy of our VARS. 
For the type of foul classification, we notice a significant improvement in accuracy with increasing dataset size, especially at the beginning. 
However, the accuracy reached a plateau between $40$\% and $80$\% of the data.
Interestingly, we observed a sudden increase in accuracy when we increased the dataset size from $80$\% to $100$\%. 
\remove{The reason for this behaviour may come from the dataset's unbalance.}
\add{This may be attributable to our unbalanced dataset.}
The dataset contains numerous ``Standing tacklings'' and ``Tacklings'', while many of the other labels are underrepresented. 
Increasing the dataset size from $40$\% to $80$\% may not have improved accuracy if the model still struggles to generalize to certain actions due to a limited number of training samples.
However, increasing the dataset size to $100$\% could have provided the model with the additional data necessary to better generalize actions.
Moreover, Figure~\ref{fig:sensitivityAnalysis} reveals that our VARS is significantly more prone to epistemic uncertainty for smaller datasets, as indicated by the high standard deviation. 

In contrast, the offense severity curve in Figure~\ref{fig:sensitivityAnalysis} initially shows a sharp increase, but later demonstrates a slower growth.
Yet, with each increase in the dataset size, the accuracy improves, which confirms that more data would further improve the performance.
The reason for this lies in the significant variability in the visual appearance of an offense with ``No card'', ``Yellow card'', or ``Red card''. For instance, a yellow card can be the outcome of a \remove{tackling} \add{tackle}, or it can be the result of a player holding an opponent's shirt. 
Although both instances may result in a yellow card, their visual representations differ significantly.
To accurately determine whether an action is an offense or not and the corresponding severity, the model needs plenty of examples to learn the underlying distribution.

\begin{figure}[ht]
    \centering
    \includegraphics[width=1\linewidth]{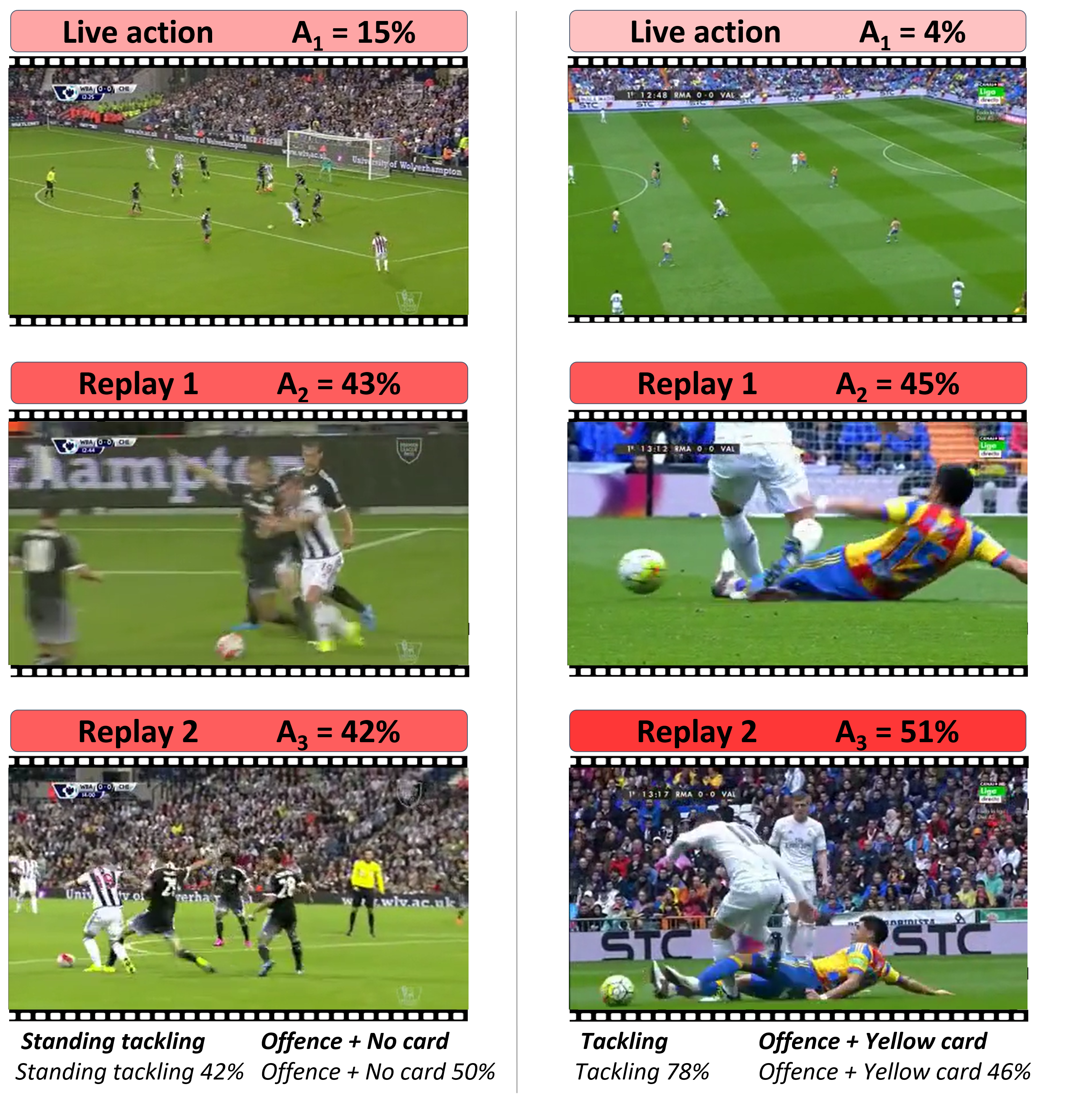}
    \caption{\textbf{Qualitative results.} VARS prediction on two examples where the attention score of each view is given in percentage. The ground truth is given in bold and the model prediction with the confidence is given in italic. 
    }
    \label{fig:qualitativeres}
\end{figure}

\mysection{Qualitative results.}
Figure~\ref{fig:qualitativeres} shows the prediction of our VARS on two examples with a 3-view setup. 
In both examples, the VARS correctly determines the type of foul and correctly classifies both actions as a foul with the correct severity.
Furthermore, the attention scores offer valuable insights into the contribution of different views or camera angles to the decision-making process of the model.  
In both cases, the ``live action clips'' have the lowest attention score, confirming our intuition that they were filmed from too far away to make an accurate decision.
Both replays have a similar attention score, as they both offer a lot of information to the model. 
However, we can see that the most informative view has a slightly higher attention score.
The attention score provides insight on which views contribute the most to classifications and helps us better understand how the model processes the visual data. 
This interpretability is especially important when the VARS is used in practice, as it is essential for fans, players, and referees to understand the reasoning behind decisions and feel confident that the technology is improving the fairness and integrity of the sport. 
Finally, the attention scores assigned to each view can assist broadcasters in automatically selecting the optimal camera angle for broadcasting purposes.
Furthermore, it can support the VAR and helps speed up the review process by automatically proposing the most informative camera perspective.
This is particularly useful at a professional level, where the VAR can have up to 30 different camera perspectives at their disposal, making finding the optimal camera a challenge on its own.
The attention scores would provide valuable information by highlighting the views that are more likely to provide crucial details, to accelerate the decision-making process during the VAR review.

\section{Human study}\label{sec:survey}
In contrast to classical classification tasks that involve well-defined and easily separable classes, determining whether an action in football constitutes a foul may be subjective. 
Despite the definitions and regulations provided by the \textit{Laws of the Game}~\cite{IFAB2022Laws}, the rule book published by the IFAB regarding when an action in football is considered a foul and its corresponding severity, these guidelines are still open to interpretation, leading to differing opinions about the same action.
In practice, many actions fall into this gray area where both interpretations, foul or no foul, could be considered correct. 
In this study, we first analyze whether and how the performance of our VARS aligns with human performance (\ie, referees and football players) by comparing the accuracy of the type of foul and offense severity classifications between VARS and our human participants.
Secondly, we conduct an inter-rater agreement analysis of human decisions to quantify the extent of agreement among our human participants.

\mysection{Experimental setup.}
The study involves two distinct groups of participants with different expertise in football: ``Players'' and ``Referees''.
The first group consisted of 15 male individuals aged 18 or older (with a mean M = $23.06$ and a standard deviation SD = $3.49$ years), who had been playing football for a minimum of three years (M = $8.71$ and SD = $3.32$ years).
The second group consisted of $15$ male individuals aged $18$ or older (M = $25.33$ and SD = $4.51$ years), who are certified football referees and have officiated in at least $200$ official games (from $223$ to $1150$ games).
Both groups analyzed $77$ actions, each presented with three different camera perspectives simultaneously. 
The participants could review the clips several times and watch the actions in slow motion or frame-by-frame, without any time restriction. To reduce bias, the actions were shown in a different random order to each participant.
For each action, we measured the time taken by the participants to make their decision. This time was measured from the moment the participants started the video until they clicked on the `Next video' button.
For each action, the participants had the same classification task as presented in Section~\ref{ExperimentTask}. Specifically, they had to determine the type of foul, if the action was a foul or not, and the corresponding severity. For each action, we use the annotations from the SoccerNet-MVFoul dataset as ground truth to determine the accuracy for each participant. 
An important note is that the participants have a clear advantage over our VARS as they view clips lasting 5 seconds, with a frame rate of $25$ fps, while our model gets a 1-second clip at $16$ fps as input.
Finally, let us note that our study was approved by the local university’s ethics committee (2223-080/5624). 
All analyses were performed using the JASP software.

\subsection{Comparison to human performance}
Table~\ref{tab:refereeVSplayerVSvars} shows the average accuracy compared to the ground truth of players, referees, and our VARS, respectively.
These results align with similar studies \cite{MacMahon_2007, Spitz_2016, Pizzera2022TheVideo}, where the referees had an overall decision accuracy ranging from 45\% to 80\%.

In terms of the type of foul categorization, players (M = $0.752$, SD = $0.055$) were numerically more accurate than referees (M = $0.704$, SD = $0.120$), but this difference was not statistically significant, as shown by an independent samples Student t-test, \textit{t}(28) = $1.421$, \textit{p} = $0.166$, \textit{d} = $0.519$, 95\% CI = [$-0.214$ - $1.243$]. Mean confidence levels in these categorizations were comparable between players (M = $3.64$, SD = $0.28$) and referees (M = $3.71$, SD = $0.32$), \textit{t}(28) $<$ 1.

\begin{table}[t]
    \centering
    \resizebox{\linewidth}{!}{
    \begin{tabular}{l|lc|lc|c}
         &  \multicolumn{2}{c|}{\bf Type of Foul} & 
        \multicolumn{2}{c}{\bf Offence Severity} & \bf Time \\ \midrule
          & \bf Acc. & \bf Conf. & \bf  Acc. & \bf Conf. & \\ 
       Players  &\bf75\%     &3.6   &58\%   &3.3 &$41.53$\\ 
       Referees     &70\%  &3.7   &\bf60\% &3.6 &$38.01$\\
       VARS     &60\%     &-   &51\%     &- & \bf 0.12 \\ 

    \end{tabular}
    }
    \caption{ \textbf{Accuracy comparison between referees, players, and our VARS.} The survey was performed on a subset of the test set of size 77. The time is given in seconds and represents the average time needed to make a decision. Acc. stands for accuracy and conf. for confidence. A rating of 5 indicates high confidence, while a rating of 1 indicates low confidence.
    }
    \label{tab:refereeVSplayerVSvars}
\end{table}

As for determining if an action corresponds to a foul and the corresponding severity, referees were slightly more accurate (M = $0.594$, SD = $0.091$) than players (M = $0.582$, SD = $0.061$). However, this difference was not statistically significant, \textit{t}(28) = $-0,401$, \textit{p} = $0.691$, \textit{d} = $-0.147$, 95\% CI = [$-0.862$ - $0.571$]. Although the accuracy of players and referees was comparable, referees were more confident in their severity judgments (M = $3.67$, SD = $0.36$) than players (M = $3.33$, SD = $0.39$), \textit{t}(28) = $-2.3$, \textit{p} = $0.029$, \textit{d} = $-0.839$, 95\% CI = [$-1.581$ - $-0.084$]. Referees' higher confidence might be due to their specific experience in assessing fouls and their severity on the field. 

Overall, our results suggest that the accuracy of players and referees is comparable. The Bayesian version of the Student t-test provides support for this null hypothesis with Bayes factors BF10 of 0.732 and 0.366 for the type of foul and offense severity task, respectively.
There is a possibility that this lack of difference between groups is due to power issues, \ie, the sample size being too small. Replication studies conducted on larger groups would be valuable in revealing potential differences between the two human groups.

As we do not have a standard deviation for the VARS, we conducted two One-Sample t-tests to compare its performance against humans (players and referees were grouped as their accuracy was comparable). For action categorization, humans (M = $0.728$, SD = $0.095$) were significantly more accurate than our VARS (M = $0.597$), \textit{t}(29) = $7.556$, \textit{p} $<$ .001, \textit{d} = $1.379$, 95\% CI = [$0.870$ - $1.876$]. Humans were also more accurate (M = $0.588$, SD = $0.081$) than our VARS (M = $0.508$) for offense severity judgments, \textit{t}(29) = $5.492$, \textit{p} $<$ .001, \textit{d} = $1.003$, 95\% CI = [$0.556$ - $1.437$].
This difference in performance might be due to differences in training between our VARS and humans. Players and referees have accumulated an extensive amount of experience in \soccer, through officiating, playing, and watching the game for countless hours. 
In contrast, our VARS has only been trained on an unbalanced training set of $2{,}916$ actions, where some types of labels only occur a few times. For example, there are only 27 fouls with a red card in the training set, making it difficult for the model to precisely learn the difference between a foul with a yellow card and one with a red card. 
Considering the difficulty of the task and the significant experience disadvantage of our VARS compared to humans, the current results are promising.
Further, it is notable that our VARS only requires $120 ms$ to reach a decision, which is more than $300$ times faster than humans. 
Both referees and players require a similar amount of time to make a decision. On average, players take around $41.53$ seconds and referees $38.01$ seconds, which is similar to the average time of $46$ seconds taken for the VAR to make a decision as reported by L\'opez~\cite{Lopez2023Average}.

\begin{figure*}[t]
    \centering
    \includegraphics[width=\linewidth]{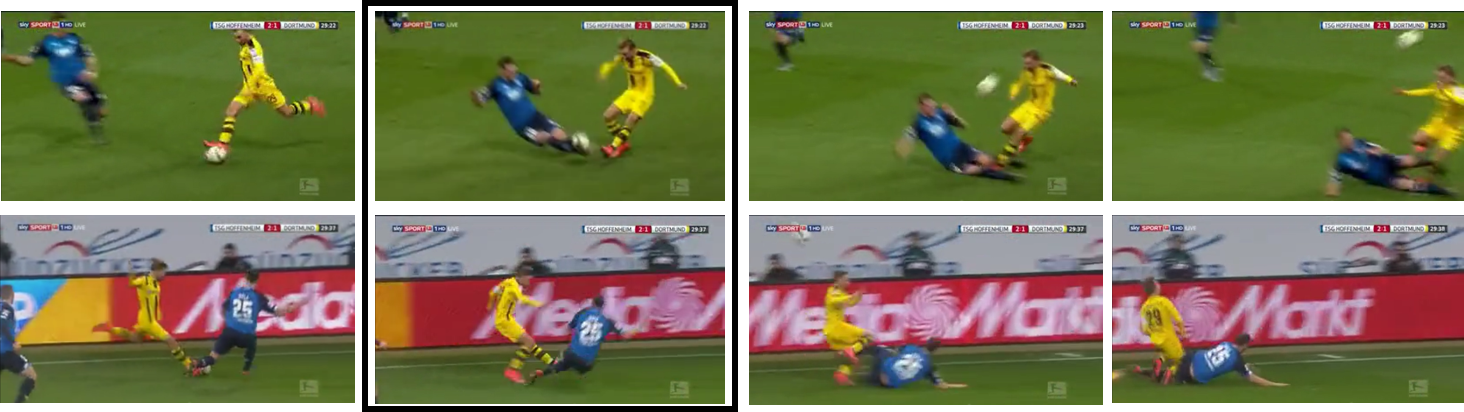}
    \caption{
    \textbf{Example of the subjectivity of human choices.}
   Decisions taken by our participants: ``No offense'', ``Offense + No card'', and ``Offense + Yellow card''.
    }
    \label{fig:examples_subjectivity}
\end{figure*}

\begin{table}
    \centering
    \resizebox{1\linewidth}{!}{
    \begin{tabular}{l|cccc}
 \textbf{Nb. of different decisions} &\textbf{1} & \textbf{2} &\textbf{3} &\textbf{4} \\\midrule
\textbf{High-level referees}    &16\% &56\% &28\% & 0\%   \\
\textbf{Referee talents}    &2\% &60\% &38\% &0\%   \\

\bottomrule
    \end{tabular}
    }
    \caption{ \textbf{Similarity analysis of the results for Offense Severity classification.} Among high-level referees, $28\%$ of cases result in three different decisions being made for the same action. For referee talents, this percentage even increases to $38\%$. These results show the significant challenge involved in determining whether an action should be classified as a foul and assessing its corresponding severity. 
    }
    \label{tab:refereeanalysis}
\end{table}

\subsection{Inter-rater agreement}
In this subsection, we investigate the reliability and consistency of humans in determining whether an action constitutes a foul and its severity.
To assess the level of consensus among humans, we calculated inter-rater agreement in each group for the severity classification task. 
Since determining if an action is a foul and assessing its severity is the most important task, we only focus on evaluating inter-rater agreement for this aspect.
To quantify the inter-rater agreement, we calculated the unweighted average Cohen's kappa, which measures the agreement between multiple individuals. The referees achieved an unweighted average Cohen's kappa of $0.213$, indicating weak agreement. Similarly, players' agreement was weak, with a score of $0.223$. This suggests limited consistency among both groups in their assessments.
Among our 15 referees, 7 are officiating at a high level (in the highest league of their country). These referees are called ``high-level referees'' in the following. All other referees are called ``referee talents''. 
Table~\ref{tab:refereeanalysis} shows the consensus in each subgroup for the offense severity classification task. 
As can be seen, high-level and referee talents reached a consensus between themselves for only 16\% and 2\% of the actions, respectively.
In the majority of cases, multiple decisions were made for the same action, indicating the difficulty in determining whether an action should be classified as a foul and assessing its severity. Particularly among referee talents, $38$\% of actions resulted in three different decisions (out of four possible decisions to take) for the same action. 
Figure~\ref{fig:examples_subjectivity} shows an example of an action where all three decisions ``No offense'', ``Offense + No card'' and ``Offense + Yellow card'' were taken among the referees.
For certain referees, the fact that the defender plays the ball is considered enough to not award a free-kick in this situation.
However, other referees believe that even if the defender plays the ball, he disregards the danger to, or consequences for, an opponent and awards a yellow card. 
These findings underscore the complexity and subjectivity inherent in refereeing decisions, highlighting the potential for further research to improve consistency and fairness in officiating.

\section{Conclusion}
Distinguishing between a foul and no foul and determining its severity is a complex and subjective task that relies entirely on the interpretation of the \textit{Laws of the Game}~\cite{IFAB2022Laws} by each individual. Despite the challenges posed by this complex task and an unbalanced training dataset, our solution demonstrates promising results. While we have not reached human-level performance yet, we believe that VARS holds the potential to assist and support referees across all levels of professionalism in the future.

\mysection{Acknowledgement}
This work was partly supported by the King Abdullah University of Science and Technology (KAUST) Office of Sponsored Research through the Visual Computing Center (VCC) funding and the SDAIA-KAUST Center of Excellence in Data Science and Artificial Intelligence (SDAIA-KAUST AI). 
J. Held and A. Cioppa are funded by the F.R.S.-FNRS. The present research benefited from computational resources made available on Lucia, the Tier-1 supercomputer of the Walloon Region, infrastructure funded by the Walloon Region under the grant agreement n°1910247.

\section{Declarations}
\mysection{Availability of data and code.} The data and
code are available at these addresses \url{https://github.com/SoccerNet/sn-mvfoul}

\mysection{Conflict of interest.} The authors declare no conflict of interest.

\mysection{Open access.}



\bibliography{bib/abbreviation-short,
bib/abbreviation-this-paper,
bib/action,
bib/activity,
bib/dataset,
bib/labo,
bib/learning,
bib/multiview,
bib/referee,
bib/soccer,
bib/sports}


\end{document}